\documentclass[11pt,a4paper]{article}

\usepackage[T1]{fontenc}
\usepackage[utf8]{inputenc}
\usepackage{lmodern}
\usepackage[margin=2.5cm]{geometry}
\usepackage{amsmath,amssymb}
\usepackage{graphicx}
\usepackage{booktabs}
\usepackage{threeparttable}
\usepackage{caption}
\usepackage{authblk}
\usepackage[numbers,sort&compress]{natbib}
\usepackage[colorlinks=true,linkcolor=blue,citecolor=blue,urlcolor=blue]{hyperref}
\usepackage{microtype}

\graphicspath{{figures/}}
\title{EEG-Based Classification of Hypnotic Susceptibility Using Two-Dimensional
Convolutional Neural Networks}

\author[1,*]{Krzysztof Wrobel}
\author[1]{Grzegorz M.\ Wojcik}
\author[1]{Filip Postepski}
\author[1]{Andrzej Kawiak}
\author[2]{Katarzyna Zemla}
\author[2]{Grzegorz Sedek}

\affil[1]{Department of Neuroinformatics and Biomedical Engineering, Institute of Computer
Science, Maria Curie-Sk\l{}odowska University, Akademicka~9, 20-031 Lublin, Poland}
\affil[2]{Institute of Psychology, SWPS University, Chodakowska~19/31, 03-815 Warsaw, Poland}
\affil[*]{Corresponding author: \href{mailto:krzysztof.wrobel@mail.umcs.pl}{krzysztof.wrobel@mail.umcs.pl}}

\date{}

\begin{document}

\maketitle

\begin{abstract}
\noindent\textbf{Background/Objectives:} Guided Imagery (GI) is a mind-based relaxation
technique widely used in clinical and therapeutic settings. Objectively assessing whether a
patient has entered a relaxed state remains a challenge for clinicians. This study investigates
electroencephalography (EEG)-based classification of hypnotic susceptibility using
two-dimensional Convolutional Neural Networks (2D~CNNs) applied to minimally preprocessed
signals, without manual feature extraction.
\textbf{Methods:} Data were collected from 62 male participants (32 Control, 30 Hypno) using
a 256-channel dense array amplifier during Guided Imagery or cognitive task conditions. A
sliding-window framework evaluated up to 868 independent models across seven anatomically
defined brain regions and 124 ten-second time windows. Segment-level random-split screening
was followed by Leave-One-Subject-Out (LOSO) cross-validation to obtain subject-independent
performance estimates, and an ensemble approach combined predictions across $K = 10$ windows.
\textbf{Results:} The best single-window model achieved accuracy of 0.82 on the random split.
Under LOSO, anterior regions generalised better across subjects: the frontal region achieved
mean accuracy of 0.646 (AUC: 0.692). An ensemble of $K = 10$ windows further improved the
prefrontal region to mean accuracy of 0.738. The occipital region, dominant in random-split
screening, collapsed to near-chance under LOSO, revealing subject-specific EEG fingerprinting
rather than genuine group differences.
\textbf{Conclusions:} Bandpass filtering and z-score normalisation suffice for competitive EEG
classification without manual feature extraction. Anterior cortical regions are identified as
the most generalisable neural markers of hypnotic susceptibility.

\end{abstract}

\section{Introduction}

Relaxation techniques have demonstrated considerable efficacy in supporting patients suffering from
various medical conditions and mental health disorders. Studies indicate that oncological patients
who incorporated relaxation practices into their care routines exhibited improved responses to
clinical treatment. Consequently, further development and promotion of relaxation-based interventions
may represent a meaningful strategy for enhancing overall patient quality of life.

Guided Imagery (GI) is a mind-based therapeutic approach that leverages the imaginative capacity of
an individual to induce changes across physical, emotional, and spiritual dimensions~\cite{fitzgerald20095imagery}.
Widely employed within psychotherapeutic practice, this method integrates relaxation strategies with
the deliberate construction of mental imagery engaging all five sensory modalities: vision, audition,
touch, taste, and olfaction. The fundamental objective of this technique is to consciously evoke
particular mental representations capable of modifying both physiological responses and emotional
states through the use of imagination~\cite{la2011content}. As a relaxation intervention, GI has been extensively
utilized and demonstrated to be effective in alleviating test anxiety and managing stress arising from
diverse sources~\cite{relaxPreg,stephens1992imagery,nguyen2018nature}. A study investigating the reduction of state anxiety conducted
by Nguyen et al.~\cite{nguyen2018nature} employed both nature-based visualizations (e.g., forest environments) and
non-nature-based settings (e.g., urban and office surroundings). Nature-oriented imagery was found
to yield superior anxiolytic effects. Furthermore, the application of Guided Imagery in the context
of chronic pain conditions~\cite{baird2004pilot} has shown notable benefits, as soothing mental images may serve
to interrupt nociceptive signaling. Given that GI is also employed to alleviate cancer-related pain,
investigating its broader influence on patients' psychological and physiological states warrants
further attention~\cite{roffe2005systematic}. Additionally, emerging evidence suggests that Guided Imagery holds promise
in the management of sleep disorders, with the visualization of serene environments --- such as a
tranquil beach or forest setting --- shown to attenuate both physical and cognitive arousal that
impedes sleep onset~\cite{morin2007insomnia}.

Guided Imagery employs mental visualization to elicit sensory experiences and is widely regarded as
one of the oldest known healing modalities. It is conceptualized as an internal perception of an
event in the absence of actual external stimuli, encompassing both sensory and cognitive
dimensions~\cite{heinschel2002descriptive}.
Guided Imagery and meditation share notable similarities, as both incorporate relaxation techniques
and mental visualization to influence physical and emotional states. These practices are effective
in calming the mind and body, reducing stress and anxiety, and promoting an overall sense of
well-being.
With respect to the Focused Attention (FA) and Open Monitoring (OM) forms of meditation, Guided
Imagery may integrate elements of both modalities. During a Guided Imagery session, individuals may
direct their attention toward specific mental images or scenarios (FA) while simultaneously remaining
open to the sensory impressions and emotional responses that arise throughout the visualization
process (OM). Accordingly, Guided Imagery may be conceptualized as a synthesis of FA and OM
meditation approaches, offering a structured framework for visualization while allowing for
unrestricted awareness of internal experiences~\cite{kaduson2021play}. Guided Imagery is recognized for its capacity
to influence a range of physiological systems, including the respiratory, cardiovascular, metabolic,
gastrointestinal, and immune systems. Its effects on these systems are mediated through the modulation
of hypothalamic-pituitary-adrenal axis activity, thereby facilitating a state of relaxation and
promoting overall physiological well-being~\cite{de2019effectiveness,sabatinelli2006neural}.

Electroencephalography (EEG) represents a viable methodology for assessing whether patients have
attained a state of relaxation. Scalp EEG is a non-invasive technique employed to record the
bioelectrical activity of the human brain. In comparison to alternative neuroimaging modalities,
such as positron emission tomography (PET) or magnetic resonance imaging (MRI), it offers the
advantages of lower cost and reduced patient burden~\cite{MURPHY199751,eeg_sig_proc_and_ml}. Nevertheless, the manual analysis
of multichannel EEG signals remains a challenging and time-intensive procedure. Machine learning and
deep learning approaches have been broadly adopted for the classification of diverse data types,
ranging from image data~\cite{imageNet} to various forms of biomedical signals~\cite{rcnnClass,cheng_ecg_2021}.
The present study aims to propose an EEG signal classification framework based on two-dimensional
Convolutional Neural Networks (2D CNNs), utilizing minimally preprocessed raw signals subjected to
basic bandpass filtering and z-score normalization as input data.

In the classification of various EEG signal types, classical machine learning (ML) algorithms, such
as Support Vector Machines (SVM), have been widely employed~\cite{svmEEG}. In the context of altered states
of consciousness and hypnosis-related EEG research, SVM-based approaches have demonstrated
classification accuracy (ACC) of approximately 80\%~\cite{svmRelaxation}.

Among state-of-the-art classification methodologies, Convolutional Neural Networks (CNNs) have been
successfully applied to EEG signal analysis~\cite{cnnAppOh}. Moreover, traditional ML methods are
progressively being superseded by deep learning frameworks in this domain. CNNs have proven
particularly effective across a broad spectrum of EEG-related applications, including motor imagery
decoding~\cite{xu2019deep}, epileptic seizure detection~\cite{zhou2018epileptic}, emotion recognition~\cite{zhang2020investigation}, as well as
research directed toward Brain-Computer Interfaces (BCIs) employing CNN-based EEG feature
extraction~\cite{chen2023brain}, and extending even to biometric identity authentication~\cite{zhang2022eeg}.

The predominant approach to EEG signal classification involves providing the classifier with
frequency band-derived features. The EEG signal is conventionally decomposed into discrete frequency
ranges, comprising delta waves (below 4~Hz), theta waves (4--7~Hz), alpha waves (8--12~Hz), beta
waves (13--30~Hz), and gamma waves (above 30~Hz). It has been demonstrated that SVM-based classifiers
can be effectively constructed using a selected subset of these frequency bands~\cite{svmRelaxation,shengEEGclassCNNfeature}.
Such an approach, however, necessitates the computation of power across specific frequency ranges.
It would therefore be advantageous to bypass manual feature extraction and instead employ CNN-based
feature extraction directly from the raw signal --- a strategy that has already been successfully
adopted in the context of emotion recognition~\cite{chenRawEEG,YanagimotoRawEEG}. Experiments reported by Baydemir et
al. demonstrated the feasibility of classifying EEG signals reflecting low and high cognitive load
using a 1D-CNN architecture with high classification accuracy~\cite{baydemir2022classification}. Similarly, CNN-based
classification of combined fNIRS-EEG mental workload signals yielded satisfactory accuracy of
89\%~\cite{fnirs_eeg}. Nevertheless, the number of studies specifically employing 1D Convolutional Neural
Networks for binary classification of relaxation and mental workload states using raw EEG signals
remains limited, indicating a need for further investigation in this area.

In our prior work, a classical classification approach was applied to distinguish between Guided
Imagery and Mental Task experimental groups~\cite{zemla2023-glm}. The Generalized Linear Model (GLM) utilized
in that study attained 81\% accuracy when operating on a narrowly defined time segment
(779--839~s) extracted from the full recording. To achieve this level of performance, the classifier
required input comprising five EEG frequency bands --- alpha, beta, delta, theta, and gamma ---
extracted from a 60-second raw signal segment. When applied to the complete recording, however, the
model yielded a substantially higher accuracy of 90.77\%.

The objective of the study is to check whether it is possible to find the best brain areas and
electrodes for classification. In this manner, all 256 channels were available for inclusion in the
training procedure. No additional preprocessing steps or artifact removal techniques were applied,
and no manual feature extraction was performed on the signal.

\section{Materials and Methods}

The signal used in this study was collected from a group of 62 male participants aged 19--24 years.
Among them, 32 were assigned to the Control Group and 30 to the Hipno Group. The described
experiments were evaluated and approved by the Bioethical Commission of Maria Curie-Sk\l{}odowska
University. All procedures were carried out in accordance with established experimental standards and
guidelines, under the supervision of qualified psychologists. Each participant consented to the EEG
recording, was informed about the aim of the study, and provided written informed consent prior to
participation.

\subsection{Inclusion and Exclusion Criteria}

The inclusion criteria required participants to be healthy individuals between 19 and 24 years of
age, male, with short hair and fluent command of Polish. Short hair was necessary to ensure higher
EEG signal quality; therefore, participants were instructed not to wash their hair or apply any
conditioners on the day of the examination, with the latest wash permitted no later than one day
prior to the study. Eligible individuals had no history of chronic illnesses, were not using
prescribed or recreational drugs, and were able to attend study sessions without any specific
technological constraints. Moreover, they were asked to refrain from consuming alcohol or taking any
medications for at least 72 hours before the experiment.

Exclusion criteria included being under 19 or over 24 years old, having long hair, insufficient
proficiency in Polish, the presence of serious or chronic medical conditions, current use of
medication or psychoactive substances, recent medical procedures, or inability to attend the
scheduled sessions. Individuals who did not meet the inclusion requirements or who reported serious
conditions, including mental health disorders, were excluded from the study. Before participation,
all subjects received information about the EEG procedures and technology and provided informed
consent to take part in the research.

Several factors motivated the recruitment of participants aged 19--24 and exclusively male subjects.
First, individuals within this age group are predominantly university students, especially those
enrolled in undergraduate and graduate programs. Second, the Institute of Computer Science is
characterized by a clear predominance of male students, which made it difficult to establish both
experimental and control groups that included female participants.

Additionally, it was observed that a large proportion of female computer science students had long
hair, which posed practical challenges for EEG data acquisition. Previous research has also reported
differences in electroencephalographic patterns between males and
females~\cite{wada1994gender,cantillo2017gender}. To maintain greater homogeneity within the
sample, the study ultimately focused solely on male participants. As a result, the conclusions drawn
apply only to men, which we acknowledge as an important limitation of this study.

All participants provided written informed consent prior to the study. One half of the sample was
assigned to the Hipno Group and listened to a Guided Imagery relaxation recording developed by a
psychologist. The remaining participants formed the Control Group and were instructed to mentally
retrieve specific categories of information, such as the names of Zodiac signs, Polish voivodships,
U.S. states, and European capitals. In both conditions, the instructions were delivered through
recordings prepared by the same psychologist.

Following each task instruction, a silent interval was included to allow participants time to think.
The Hipno Group was expected to remain relaxed throughout the session, whereas the Control Group was
deliberately exposed to cognitive effort. At the start of the experiment, members of the Control
Group were informed that they would be required to write down all recalled information after the
session, thereby increasing task engagement. Each task was time-limited. The assumption that this
procedure would elicit mental workload was supported by previous findings indicating that task
difficulty and time constraints contribute to increased cognitive load~\cite{galy2012relationship}.

Both the Guided Imagery and mental task recordings lasted 20 minutes. Participants were instructed
to keep their eyes closed, and each session was conducted with subjects lying down in a darkened
room to minimize muscle artifacts, electrical interference, and external distractions that could
affect the EEG recordings.

\begin{figure}[!h]
    \centering
    \includegraphics[width=0.9\linewidth]{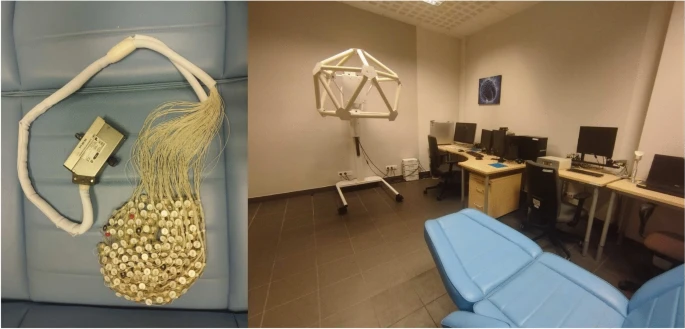}\\[6pt]
    \includegraphics[width=0.6\linewidth]{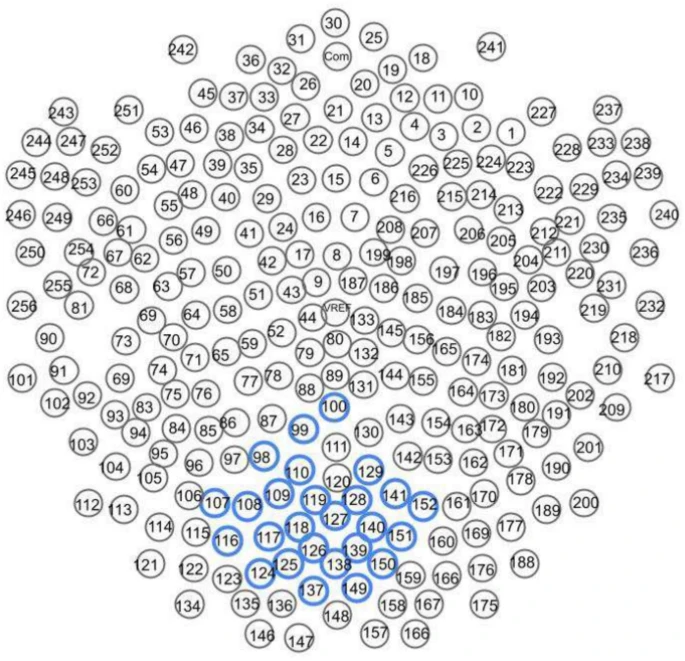}
    \caption{Top: overview of the EEG Laboratory at UMCS, Lublin, Poland.
    Bottom: EGI 256-channel EEG cap.}
    \label{fig:lab}
\end{figure}

The experiments were conducted in the EEG Laboratory of the Department of Neuroinformatics and
Biomedical Engineering of Maria Curie-Sk\l{}odowska University (UMCS) in Lublin,
Poland (Figure~\ref{fig:lab}). All trial signals were recorded at the sampling frequency of 250~Hz
with the use of a 256-channel dense array EGI GSN 130 series cap (Figure~\ref{fig:lab}). For signal
acquisition, the EGI Net Station 4.5.4 software was used.

Our dense array amplifier recorded the signal from all 256 electrodes. The objective of the study
is to identify the cortical regions and corresponding electrode subsets that exhibit the greatest
sensitivity to hypnosis-related neural patterns.

\subsection{Signal Preprocessing and Data Preparation}

EEG signals were preprocessed using the MNE Python library (version 1.6.0)~\cite{mnePython}.
Channels exhibiting excessive noise were identified, removed, and subsequently reconstructed through
interpolation to ensure consistent data dimensionality across all samples. Automatic detection of
bad channels was performed using the RANSAC algorithm, available within the pyprep
toolkit~\cite{appelhoff_2023_10047462pyprep}, which is itself derived from the PREP pipeline
originally developed for EEG preprocessing in MATLAB~\cite{fninf.2015.00016prep}. Although MATLAB
was not employed in this study, both environments yielded comparable outcomes. Each trial's signal
was bandpass filtered within the 1--40~Hz frequency range. Following filtering, each signal was
z-score normalised per channel:

\begin{equation}
    \hat{x} = \frac{x - \mu}{\sigma}
\end{equation}

\noindent where $\mu$ and $\sigma$ denote the channel mean and standard deviation, respectively,
computed over the entire recording segment.

To examine the entire signal, 10-second windows are isolated simultaneously for each brain region.
Such a large volume of computation would be infeasible in a standard sequential environment; the
asynchronous capabilities of Python were therefore leveraged, specifically the event-loop mechanism
and the built-in \texttt{asyncio} library combined with \texttt{ProcessPoolExecutor} for parallel
model training.

\subsection{Brain Region Electrode Selection}

Seven anatomically defined brain regions were investigated: prefrontal, frontal, temporal, central,
parietal, occipital, and facial. Each region was represented by a predefined subset of electrodes
from the 256-channel EGI cap, selected according to standard neuroanatomical
landmarks~\cite{geodesics2009geodesic}.
Table~\ref{tab:electrodes} lists the electrode subsets assigned to each region.

\begin{table}[!h]
\centering
\small
\caption{Electrode subsets per brain region in the 256-channel EGI GSN~130 cap.
Electrode identifiers follow the EGI convention (E\textit{n}); consecutive identifiers
are shown as ranges. $N$ is the number of electrodes per region.}
\label{tab:electrodes}
\begin{tabular}{lcp{10.5cm}}
\hline
\textbf{Region} & $N$ & \textbf{Electrode identifiers (sorted)} \\
\hline
Prefrontal & 22 &
E2, E3, E10--E12, E18--E20, E25--E27, E31--E34, E37--E39, E46--E48, E222 \\[3pt]
Frontal & 20 &
E4--E6, E13--E15, E21--E23, E28--E30, E35, E36, E40, E41, E214, E215, E223, E224 \\[3pt]
Temporal & 72 &
E1, E49, E54--E58, E61--E64, E67--E75, E82, E83, E91--E95, E102--E104,
E173, E178, E180, E181, E189--E195, E200--E204, E208--E213, E216--E221,
E225--E229, E232, E233, E250--E256 \\[3pt]
Central & 25 &
E7--E9, E16, E17, E24, E42--E44, E50--E52, E59, E65,
E182--E184, E186, E195--E198, E205--E207 \\[3pt]
Parietal & 41 &
E45, E53, E60, E66, E77--E81, E84--E90, E96--E101,
E129--E132, E141--E144, E152--E155, E161--E164, E170, E171, E179 \\[3pt]
Occipital & 40 &
E105--E119, E121--E128, E138--E140, E149--E151, E158--E160,
E167--E169, E175--E177, E188, E199 \\[3pt]
Facial & 30 &
E67, E73, E82, E91, E216--E219, E225--E230, E232, E233,
E235, E236, E240, E243, E245, E246, E248, E250--E256 \\
\hline
\end{tabular}
\end{table}

\subsection{CNN Architecture}

Based on prior research within this project~\cite{postepski2025}, a 2D CNN architecture was
selected over recurrent models such as LSTM. The classification operates on short, independent
50-sample sub-segments (0.2~s at 250~Hz). LSTM is designed for tasks in which the temporal order
of a sequence of segments carries meaning (e.g., predicting the next token or tracking a
trajectory). Here, each sub-segment is classified in isolation --- there is no recurrence between
segments, so LSTM's sequential memory provides no benefit.

Furthermore, the core experiment trains a separate model per 10-second window, meaning each model
operates exclusively on data within that window; no cross-window temporal dependencies are modelled.
The pipeline trains up to $124 \times 7 = 868$ individual models in parallel. An LSTM-based model
has more parameters, longer convergence time, and is more prone to vanishing gradients on short EEG
segments, making the lightweight 2D CNN better suited to this large-scale sliding-window design.

\subsubsection{Convolutional Block 1}

The first block begins with a Conv2D layer with 32 filters of size $3 \times 3$ and ReLU
activation, which learns local spatial-temporal features across neighbouring electrodes and time
points simultaneously. The feature maps are then spatially downsampled by a MaxPooling2D layer
with a $2 \times 2$ pool size, reducing spatial dimensions by half and retaining only the most
prominent activations. A Dropout layer with a rate of 0.25 is applied afterwards to prevent
co-adaptation of feature detectors during training.

\subsubsection{Convolutional Block 2}

The second block follows the same structure but doubles the number of filters to 64, allowing the
network to learn more abstract and diverse feature representations at a coarser spatial resolution.
A Conv2D layer with 64 filters of size $3 \times 3$ and ReLU activation is followed by another
MaxPooling2D layer ($2 \times 2$) and a Dropout layer with rate 0.25.

\subsubsection{Classifier Head}

After the two convolutional blocks, a Flatten layer linearises the 3D feature maps into a single
1D vector. This is passed to a Dense layer with 64 units and ReLU activation, which learns a
non-linear combination of the extracted features. A Dropout layer with a higher rate of 0.50 is
applied here to strongly regularise the dense representation. The final output Dense layer has 2
units with Softmax activation, producing a probability distribution over the two classes:
HIPNO and CONTROL.

\subsubsection{Training Configuration}

The model is compiled with the Adam optimiser and sparse categorical cross-entropy loss. Training
runs for 20 epochs with a batch size of 32. The dataset is split into 90\% training and 10\% test
sets using a fixed random seed of 42, ensuring reproducibility. A separate model is trained
independently for each combination of brain region and 10-second time window, yielding up to 868
models in total across the full experiment.

\subsubsection{Design Rationale}

Treating the EEG segment as a 2D matrix with electrodes along one axis and time samples along the
other allows the $3 \times 3$ convolutional kernels to capture joint spatial and temporal patterns
--- that is, correlated activations across neighbouring electrodes within a short time horizon.
Doubling filter depth in the second block follows standard CNN design practice, enabling the
extraction of increasingly abstract representations while progressively compressing spatial
information through pooling. The relatively high dropout rate of 0.50 in the dense layers
compensates for the limited number of training examples available per window-level model.

\begin{figure}[!h]
    \centering
    \includegraphics[width=0.9\linewidth]{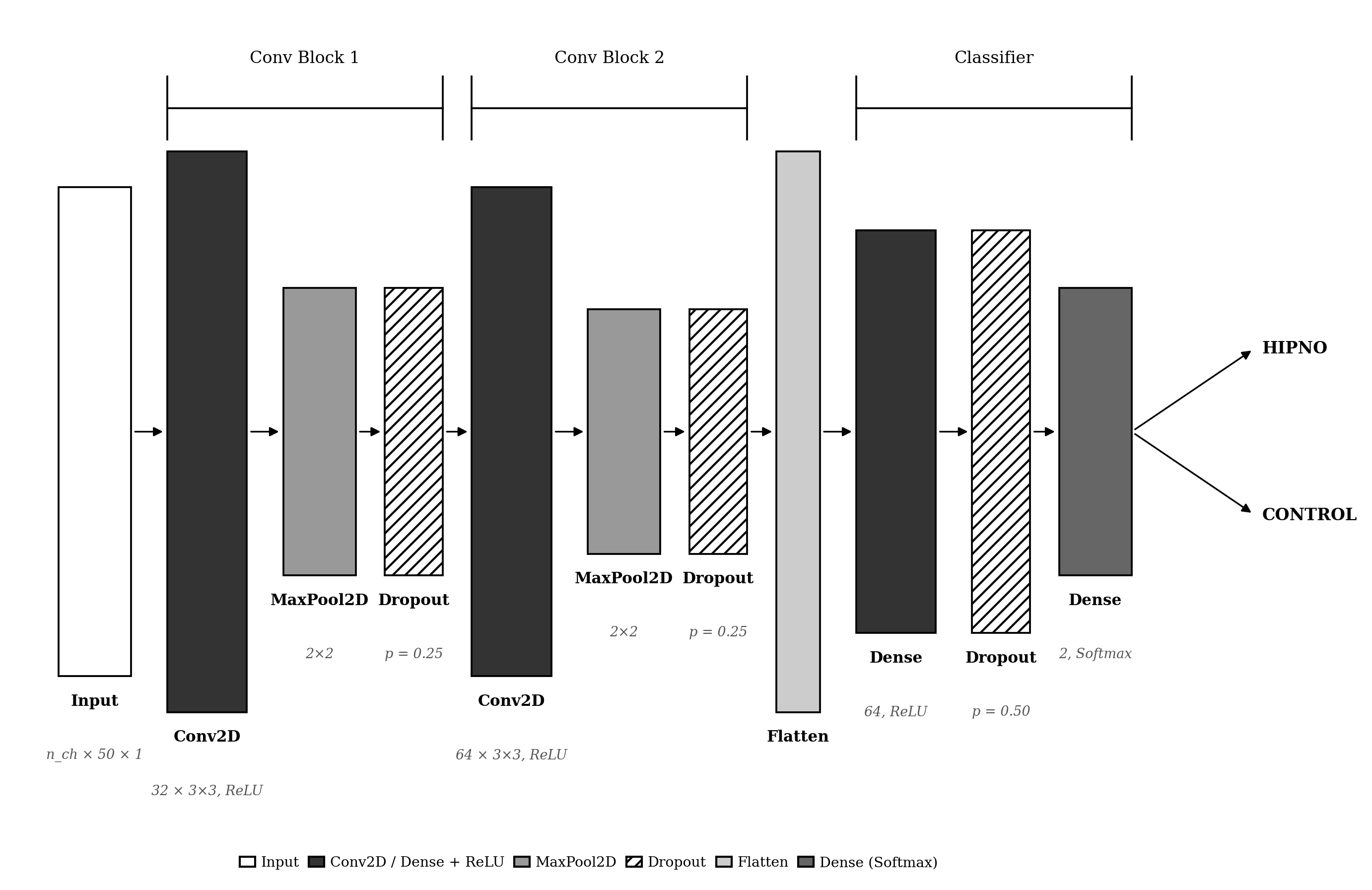}
    \caption{2D-CNN architecture used for HIPNO/CONTROL EEG classification.}
    \label{fig:cnn_architecture}
\end{figure}

\subsection{Cross-Validation Protocol}

The initial screening experiment evaluates each window--region combination using a
segment-level 90\%/10\% random train-test split with a fixed random seed of 42. Because
consecutive 50-sample sub-segments extracted from the same 10-second recording window of
the same subject may appear in both the training and test sets, this split does not
guarantee subject-independent generalisation; a classifier may learn subject-specific EEG
fingerprints rather than group-level differences. This procedure is used exclusively for
the sliding-window screening of all $124 \times 7$ window--region combinations.

To obtain a subject-independent performance estimate, Leave-One-Subject-Out (LOSO)
cross-validation was applied to the best-performing window per discriminative region.
In each of the 62 folds, the model was trained from scratch on data from 61 subjects and
evaluated on all segments of the held-out subject. Per-fold accuracy is reported as
mean~$\pm$~standard deviation across the 62 folds; aggregate F1, precision, recall, and
AUC are computed from the concatenated predictions of all folds with the HIPNO group as
the positive class.

To assess whether aggregating predictions across multiple temporal windows improves
subject-level generalisation, an ensemble variant of LOSO was additionally evaluated.
In each fold, $K$ independently trained models --- each operating on a distinct
top-ranked 10-second window of the region --- averaged their softmax probability outputs
(soft voting) to produce a single per-subject prediction. Two ensemble sizes were
compared: $K = 3$ and $K = 10$.

\subsection{Evaluation Metrics}

The primary performance metric is classification accuracy,
$\text{ACC} = (\mathrm{TP}+\mathrm{TN})\,/\,(\mathrm{TP}+\mathrm{TN}+\mathrm{FP}+\mathrm{FN})$,
where TP, TN, FP, and FN denote true positives, true negatives, false positives, and false
negatives. A window is considered \emph{discriminative} when $\text{ACC} > 0.67$.
Per-region summary statistics include the number and proportion of discriminative windows,
peak accuracy, and mean accuracy across discriminative windows.

For the best-performing window of each discriminative region, the following metrics are additionally
reported, with the HIPNO group treated as the positive class:
precision $(\text{Prec} = \mathrm{TP}/(\mathrm{TP}+\mathrm{FP}))$,
recall $(\text{Rec} = \mathrm{TP}/(\mathrm{TP}+\mathrm{FN}))$, and
F1 score $(\text{F1} = 2\,\text{Prec}\cdot\text{Rec}\,/\,(\text{Prec}+\text{Rec}))$.
Receiver Operating Characteristic (ROC) curves are computed from the model's softmax probability
outputs, and the Area Under the Curve (AUC) is reported as a threshold-independent measure of
class separability. Normalised confusion matrices are presented to characterise the distribution
of classification errors across both classes.

\section{Results}

All experiments were implemented using TensorFlow 2.18.0 and Keras 3.6.0 with Python 3.11.9.
The hardware used was an Apple MacBook Pro equipped with Apple Silicon M3 Max and 64~GB RAM,
running macOS 14 Sonoma. No hardware overclocking was applied.

The CNN model achieved classification accuracy exceeding the 0.67 threshold in five out of seven
evaluated brain regions. The occipital region demonstrated the highest proportion of discriminative
windows (112/124, 90.3\%), with a peak accuracy of 0.8026 and a mean accuracy of 0.7206, indicating
robust and consistent group separation in posterior cortex. The parietal region followed closely,
yielding 95/124 (76.6\%) discriminative windows, a peak accuracy of 0.7730, and a mean accuracy
of 0.7087.

Anterior regions showed substantially lower proportions of discriminative windows, though with
competitive peak accuracies. The prefrontal region achieved the highest single-window accuracy
across all regions (0.8224), with 24/124 (19.4\%) windows above threshold and a mean accuracy of
0.7086. The central and frontal regions produced 20/124 (16.1\%) and 13/124 (10.5\%) discriminative
windows, reaching peak accuracies of 0.8026 and 0.7829, respectively.
The temporal and facial regions yielded no windows above the accuracy threshold, indicating an
absence of discriminative EEG signal in these areas within the applied classification framework.

\begin{table}[!h]
\centering
\caption{Classification performance of the CNN model across EEG brain regions.
Discriminative windows are those exceeding the 0.67 accuracy threshold.
Total number of analysed windows per region: 124.}
\label{tab:results}
\small\setlength{\tabcolsep}{4pt}
\begin{tabular}{lcccc}
\hline
\textbf{Brain Region} & \textbf{Discriminative windows} & \textbf{Proportion (\%)} & \textbf{Peak ACC} & \textbf{Mean ACC} \\
\hline
Occipital   & 112/124 & 90.3 & 0.8026 & 0.7206 \\
Parietal    &  95/124 & 76.6 & 0.7730 & 0.7087 \\
Prefrontal  &  24/124 & 19.4 & 0.8224 & 0.7086 \\
Central     &  20/124 & 16.1 & 0.8026 & 0.6998 \\
Frontal     &  13/124 & 10.5 & 0.7829 & 0.7105 \\
Temporal    &   0/124 &  0.0 & ---    & ---    \\
Facial      &   0/124 &  0.0 & ---    & ---    \\
\hline
\end{tabular}
\end{table}

\begin{figure}[!h]
    \centering
    \includegraphics[width=\linewidth]{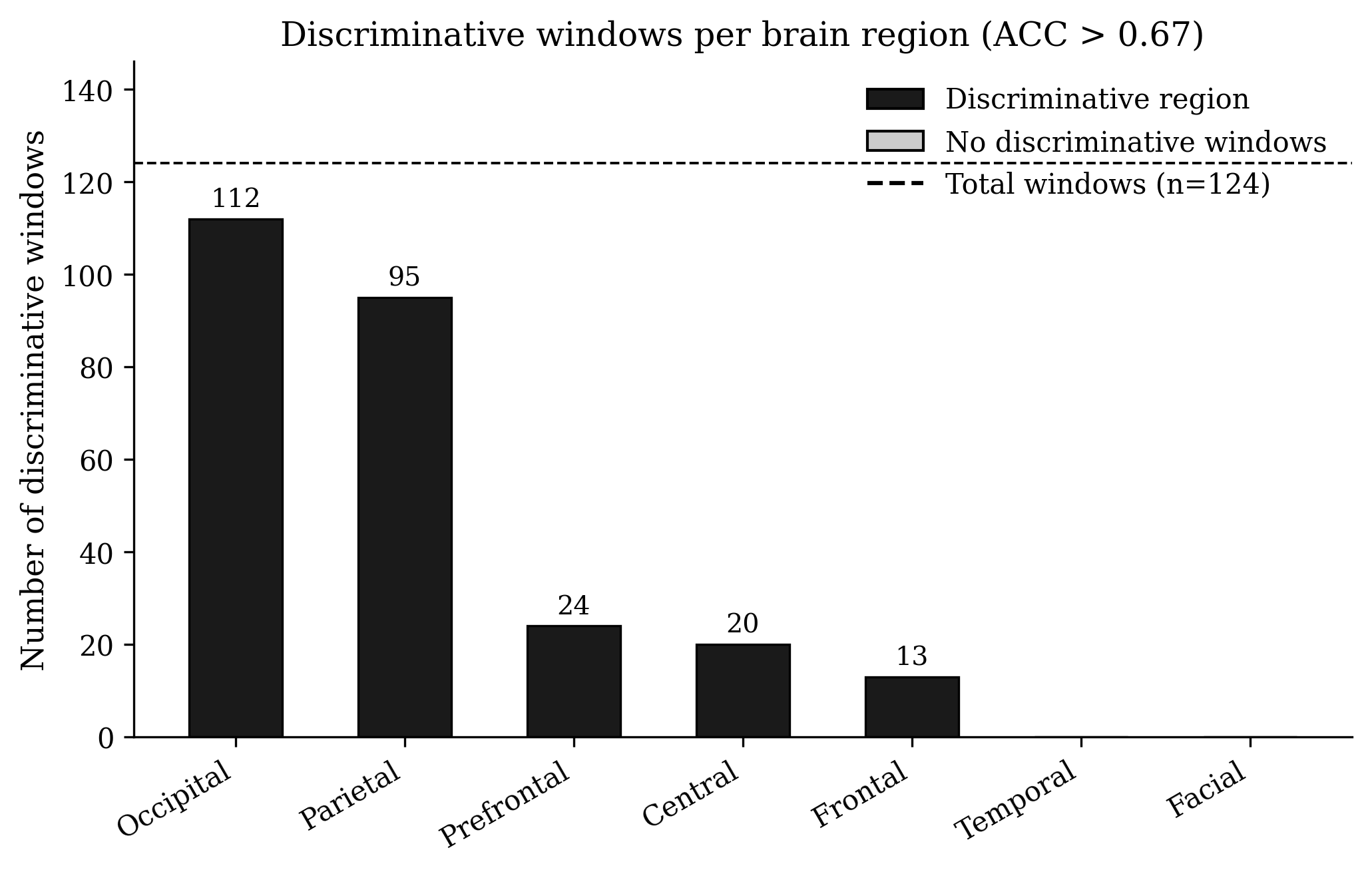}
    \caption{Number of discriminative windows (ACC~$> 0.67$) per brain region out of 124 total
    analysed windows.}
    \label{fig:discriminative_windows}
\end{figure}

\begin{figure}[!h]
    \centering
    \includegraphics[width=\linewidth]{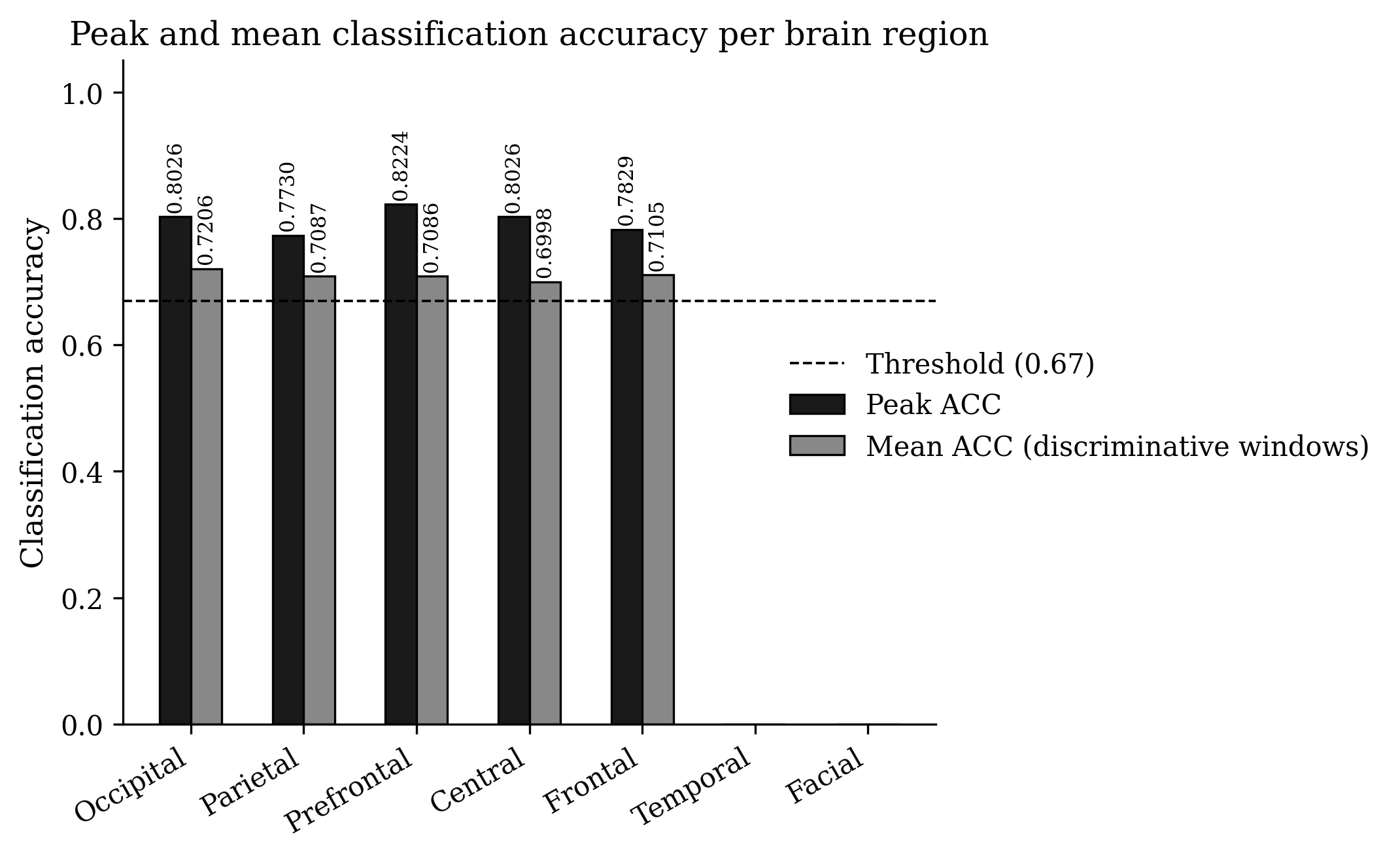}
    \caption{Peak and mean classification accuracy per brain region for regions with at least one
    discriminative window.}
    \label{fig:accuracy}
\end{figure}

\begin{figure}[!h]
    \centering
    \includegraphics[width=\linewidth]{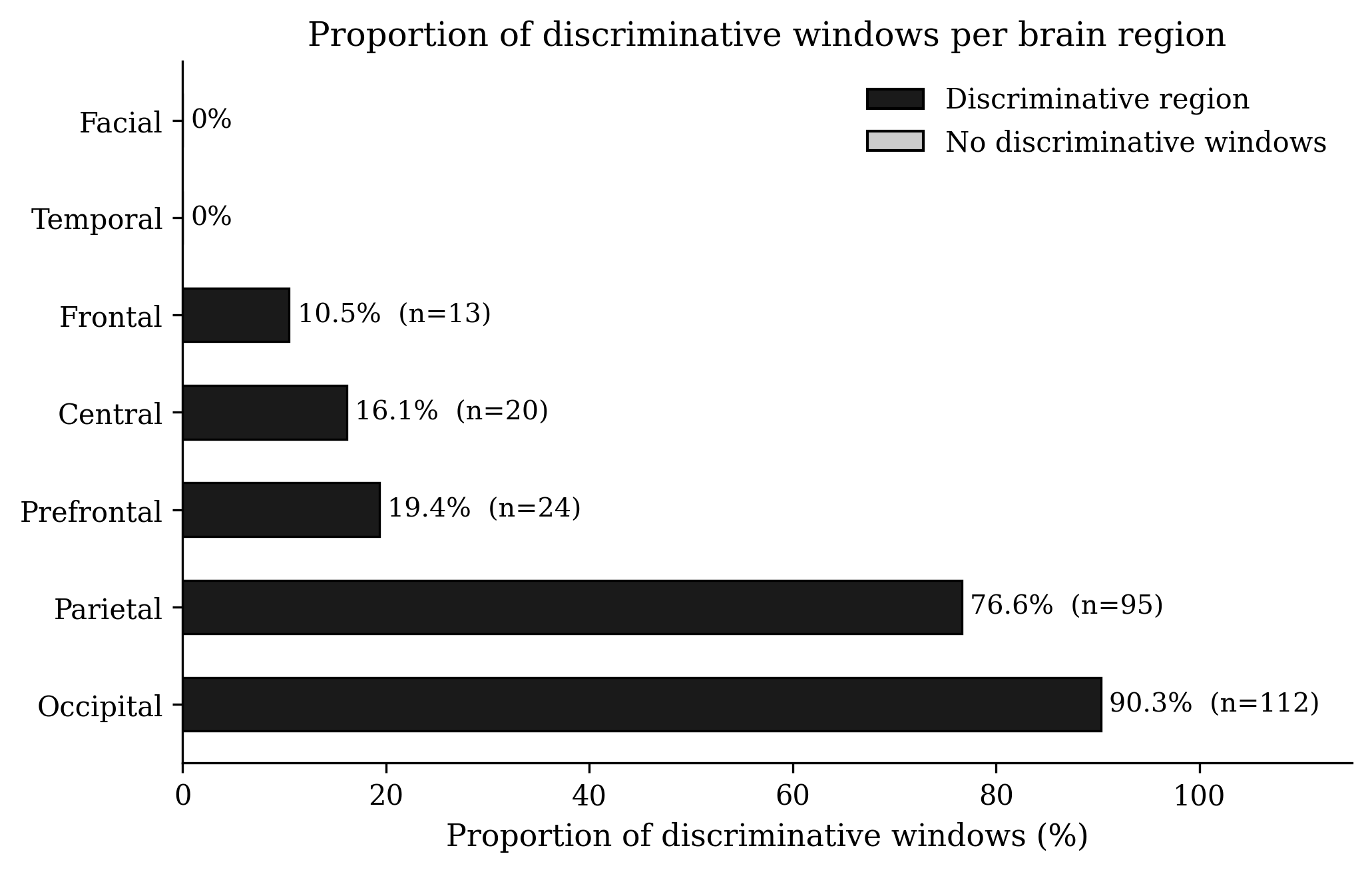}
    \caption{Proportion (\%) of discriminative windows (ACC~$> 0.67$) out of 124 total windows
    per brain region.}
    \label{fig:proportion}
\end{figure}

These findings collectively suggest a posterior-dominant pattern of EEG-based group discrimination,
in which occipital and parietal regions provide the most consistent separability between HIPNO and
CONTROL groups across time, while anterior regions --- despite occasionally reaching competitive
peak accuracies --- exhibit limited temporal stability of the discriminative signal. The absence of
classification above threshold in temporal and facial regions further delimits the spatial
specificity of hypnosis-related neural markers within the applied framework.

\begin{figure}[!h]
    \centering
    \includegraphics[width=\linewidth]{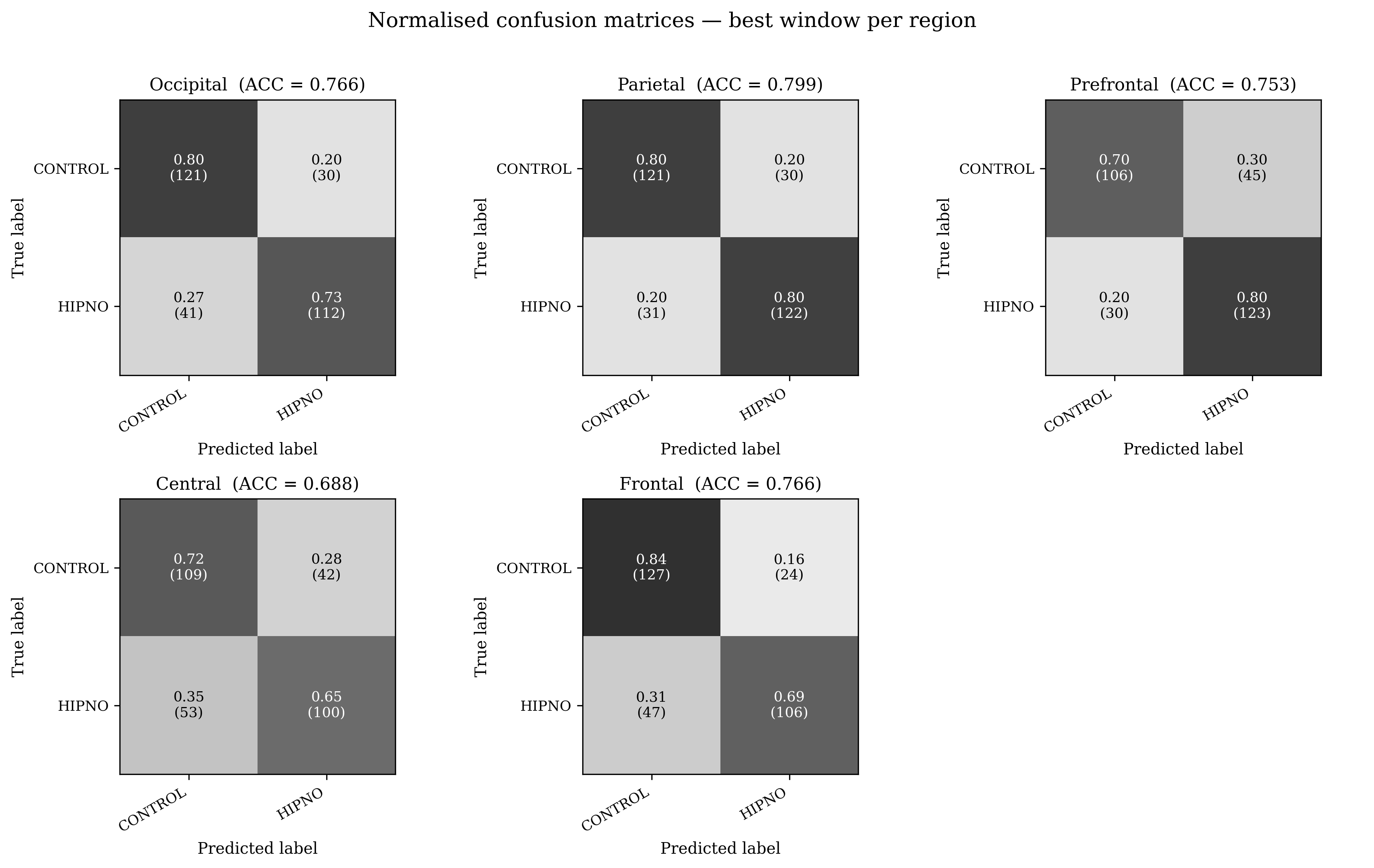}
    \caption{Normalised confusion matrices for the best-performing window per discriminative
    brain region. Values represent row-normalised proportions; raw counts are shown in
    parentheses. Rows correspond to true labels; columns to predicted labels.}
    \label{fig:confusion_matrices}
\end{figure}

\begin{figure}[!h]
    \centering
    \includegraphics[width=0.75\linewidth]{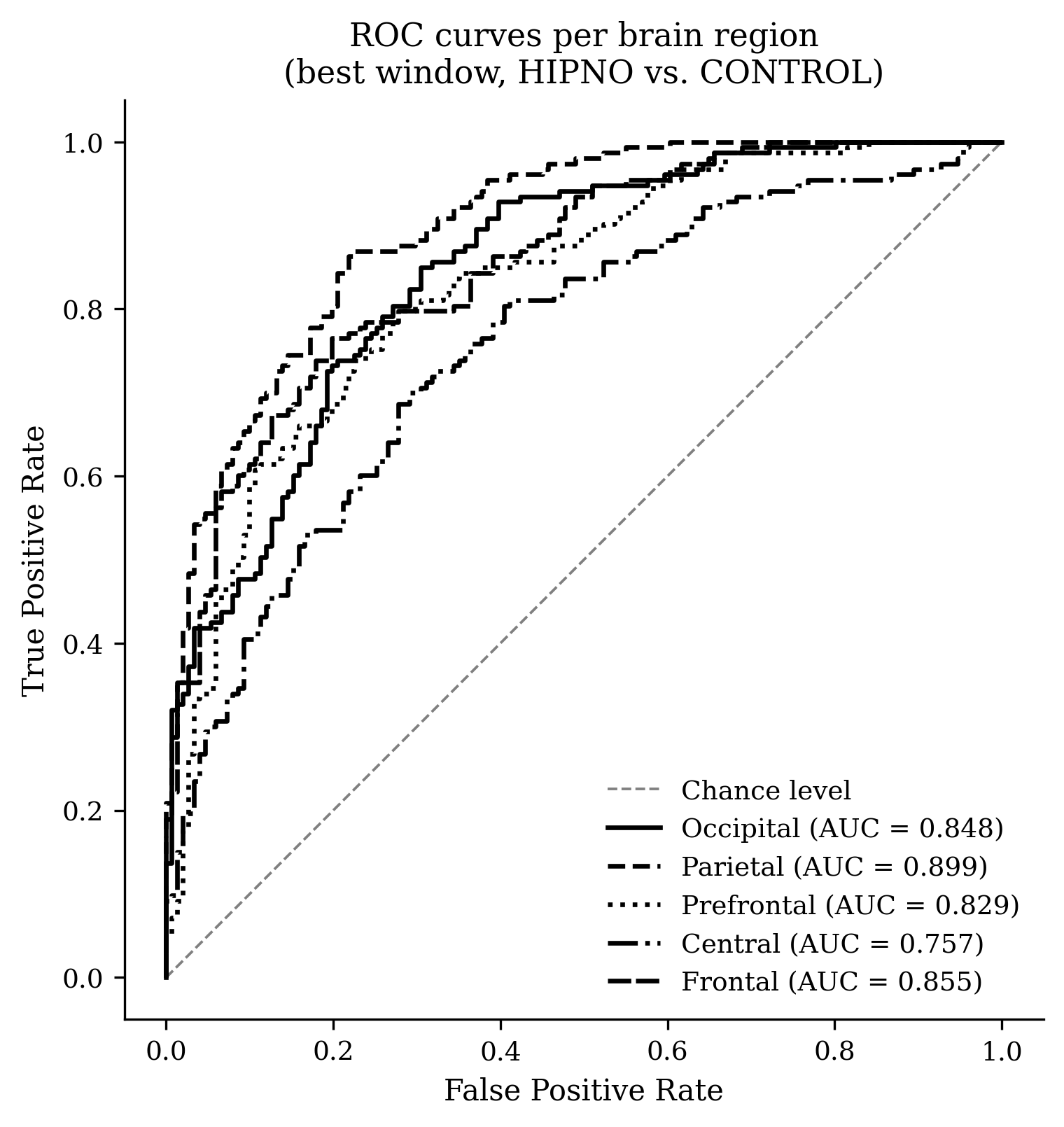}
    \caption{ROC curves for the five discriminative brain regions, computed from the
    best-performing window per region. Line styles distinguish regions for monochrome
    reproduction. AUC values are shown in the legend.}
    \label{fig:roc_curves}
\end{figure}

To characterise classification quality beyond accuracy, the best-performing window per
discriminative region was re-trained and evaluated using precision, recall, F1 score, and
ROC AUC (Table~\ref{tab:best_metrics}). Normalised confusion matrices and ROC curves are
presented in Figures~\ref{fig:confusion_matrices} and~\ref{fig:roc_curves}.

\begin{table}[!h]
\centering
\caption{Classification metrics for the best-performing window per discriminative region
(HIPNO as positive class). Values are obtained by re-training the identified best window
with a fixed random seed; run \texttt{generate\_best\_metrics.py} to reproduce.}
\label{tab:best_metrics}
\begin{tabular}{lcccccc}
\hline
\textbf{Region} & \textbf{Window (s)} & \textbf{ACC} & \textbf{F1} & \textbf{Precision} & \textbf{Recall} & \textbf{AUC} \\
\hline
Occipital  & 771--781   & 0.766 & 0.760 & 0.789 & 0.732 & 0.848 \\
Parietal   & 1121--1131 & 0.799 & 0.800 & 0.803 & 0.797 & 0.899 \\
Prefrontal & 1101--1111 & 0.753 & 0.767 & 0.732 & 0.804 & 0.829 \\
Central    & 1101--1111 & 0.688 & 0.678 & 0.704 & 0.654 & 0.757 \\
Frontal    & 1121--1131 & 0.766 & 0.749 & 0.815 & 0.693 & 0.855 \\
\hline
\end{tabular}
\end{table}

\subsection{Leave-One-Subject-Out Cross-Validation}

Leave-One-Subject-Out (LOSO) cross-validation was applied to the best-performing window
per discriminative region to provide a subject-independent generalisation estimate free
from segment-level data leakage, as described in Materials and Methods
(run \texttt{generate\_loso\_metrics.py} to reproduce).
Results across the 62 folds are summarised in Table~\ref{tab:loso}.

\begin{table}[!h]
\centering
\caption{LOSO cross-validation results for the best window per discriminative region.
ACC is mean $\pm$ std across 62 folds; F1, Precision, Recall, and AUC are computed
on concatenated fold predictions (HIPNO as positive class).}
\label{tab:loso}
\begin{tabular}{lcccccc}
\hline
\textbf{Region} & \textbf{Window (s)} & \textbf{Mean ACC $\pm$ Std} & \textbf{F1} & \textbf{Precision} & \textbf{Recall} & \textbf{AUC} \\
\hline
Occipital  & 771--781   & $0.453 \pm 0.235$ & 0.434 & 0.435 & 0.434 & 0.429 \\
Parietal   & 1121--1131 & $0.631 \pm 0.228$ & 0.627 & 0.613 & 0.641 & 0.680 \\
Prefrontal & 1101--1111 & $0.624 \pm 0.353$ & 0.616 & 0.608 & 0.624 & 0.681 \\
Central    & 1101--1111 & $0.591 \pm 0.243$ & 0.582 & 0.576 & 0.588 & 0.599 \\
Frontal    & 1121--1131 & $0.646 \pm 0.302$ & 0.638 & 0.633 & 0.641 & 0.692 \\
\hline
\end{tabular}
\end{table}

\begin{figure}[!h]
    \centering
    \includegraphics[width=0.8\linewidth]{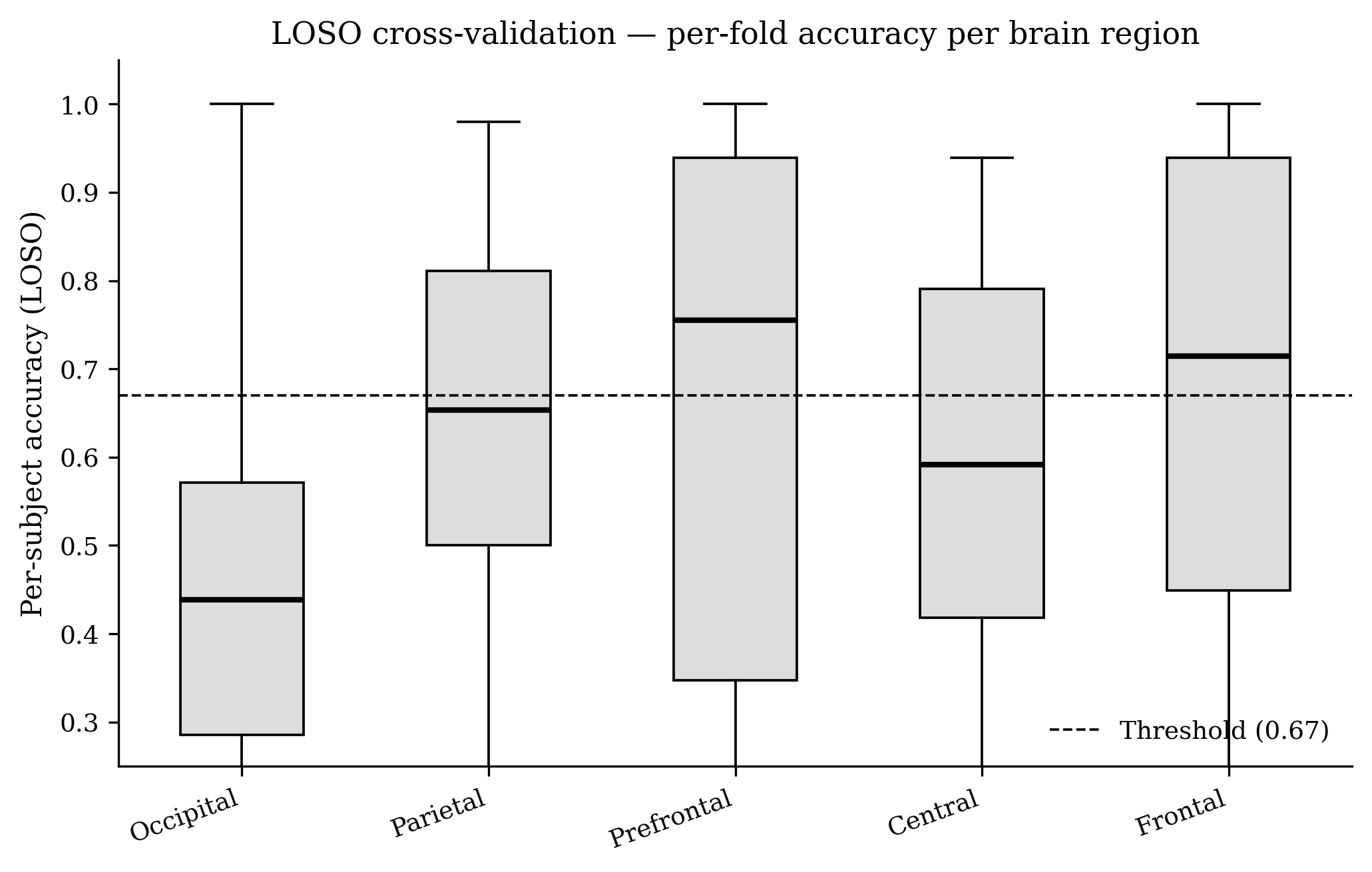}
    \caption{Per-subject accuracy distributions from LOSO cross-validation for each
    discriminative brain region. Each box represents the distribution of 62 fold
    accuracies. The dashed line marks the 0.67 discriminative threshold.}
    \label{fig:loso_accuracy}
\end{figure}

\begin{figure}[!h]
    \centering
    \includegraphics[width=\linewidth]{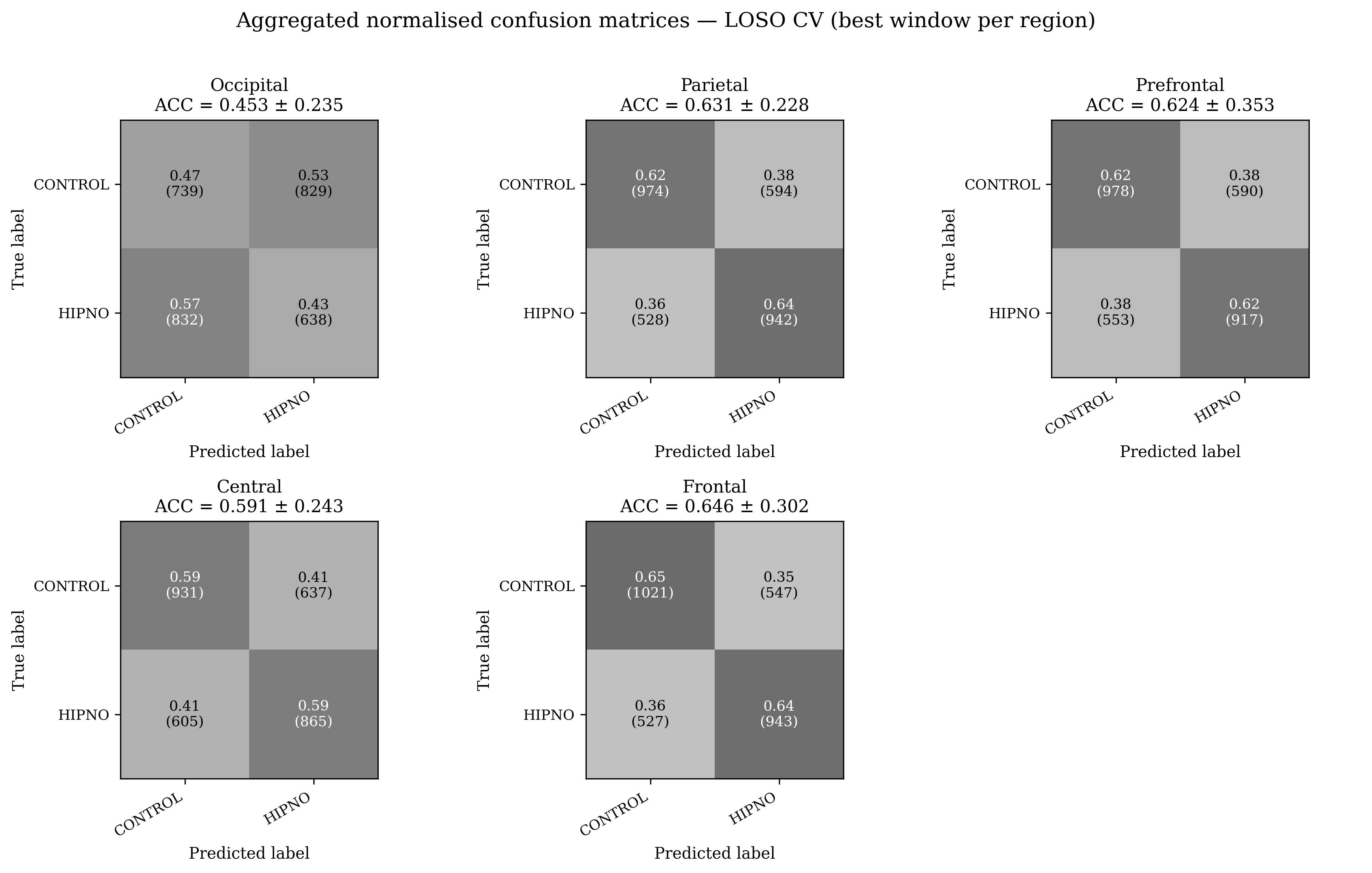}
    \caption{Aggregated normalised confusion matrices from LOSO cross-validation.
    Values are computed from the concatenated predictions of all 62 folds.
    Row-normalised proportions are shown; raw counts in parentheses.}
    \label{fig:loso_cm}
\end{figure}

\begin{figure}[!h]
    \centering
    \includegraphics[width=0.75\linewidth]{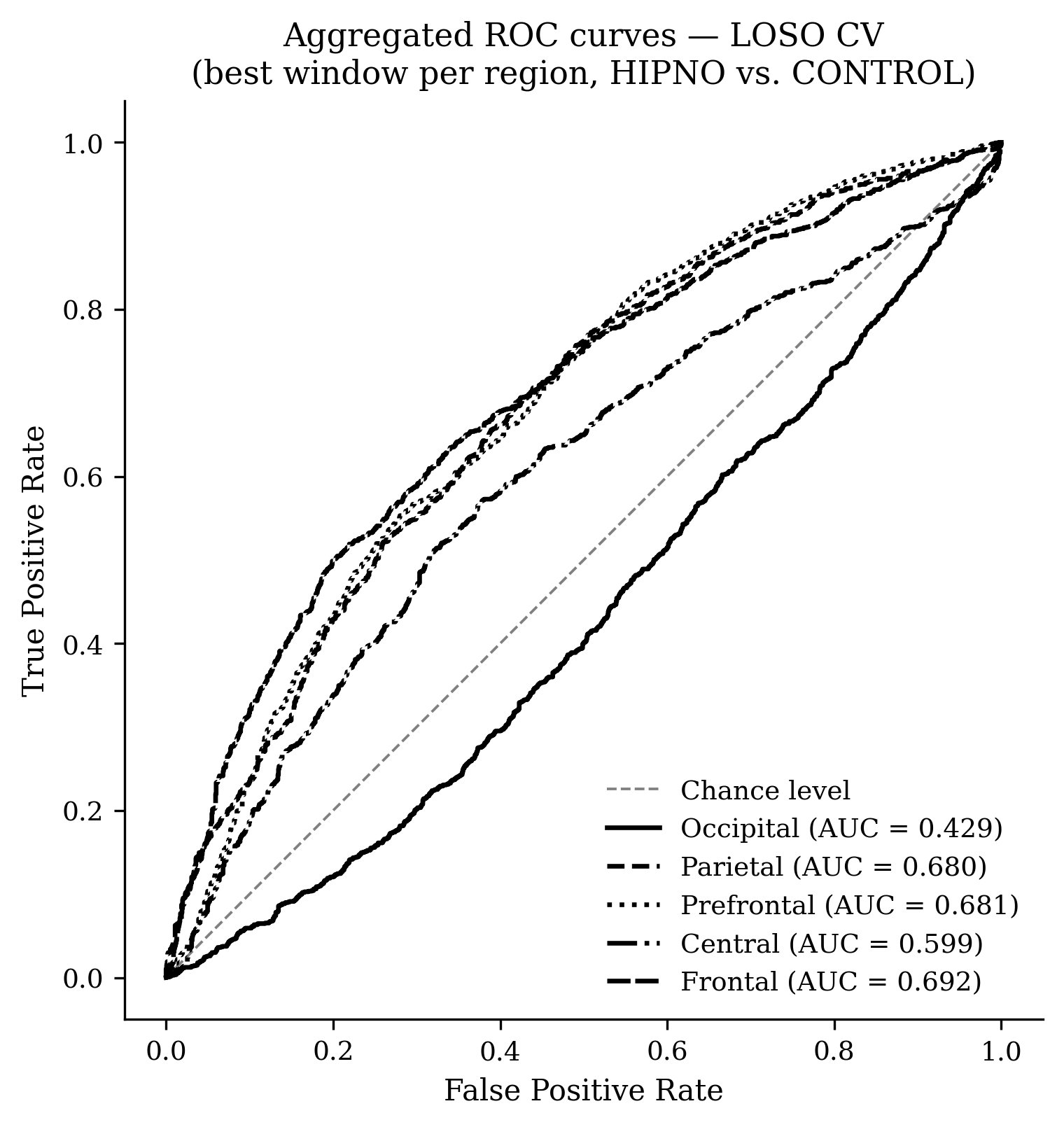}
    \caption{Aggregated ROC curves from LOSO cross-validation, computed from
    concatenated fold predictions. AUC values are shown in the legend.}
    \label{fig:loso_roc}
\end{figure}

\subsection{Ensemble LOSO Cross-Validation}

\begin{figure}[!h]
    \centering
    \includegraphics[width=0.8\linewidth]{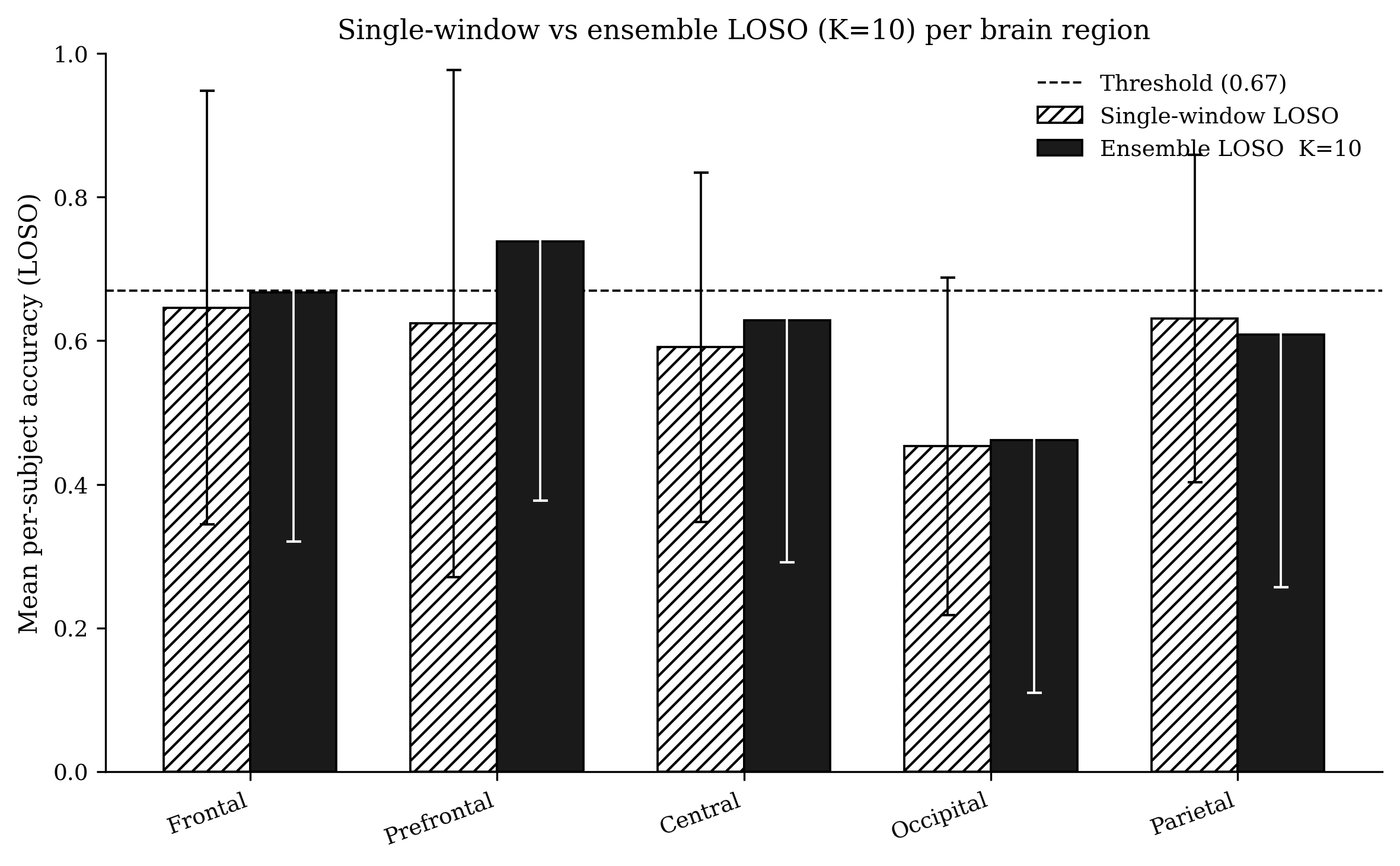}
    \caption{Per-subject accuracy distributions from ensemble LOSO cross-validation (K~=~10)
    for each discriminative brain region. The dashed line marks the 0.67 discriminative
    threshold.}
    \label{fig:ensemble_loso_accuracy}
\end{figure}

\begin{figure}[!h]
    \centering
    \includegraphics[width=\linewidth]{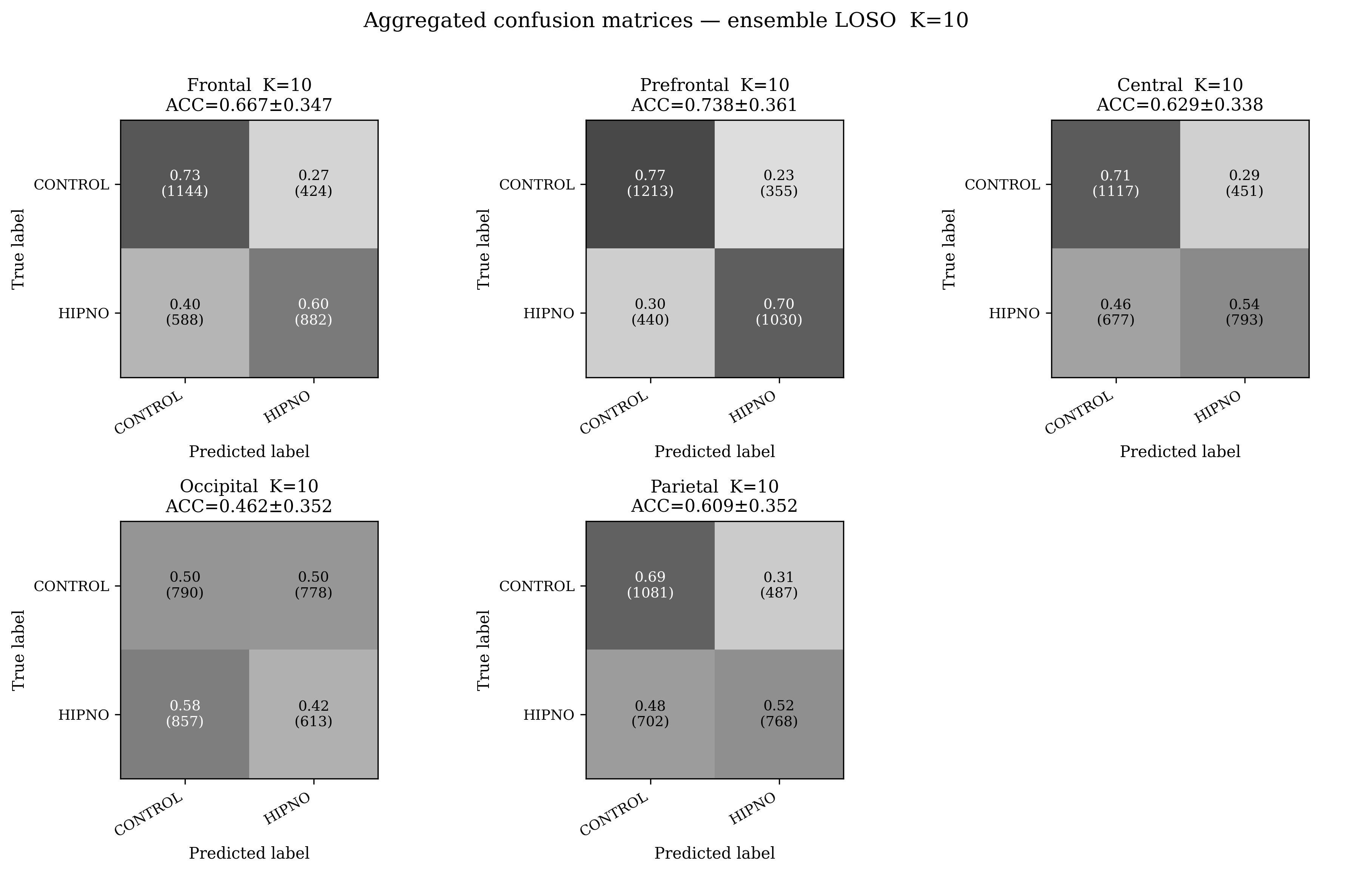}
    \caption{Aggregated normalised confusion matrices from ensemble LOSO cross-validation
    (K~=~10), computed from concatenated predictions across all 62 folds.
    Row-normalised proportions are shown; raw counts in parentheses.}
    \label{fig:ensemble_loso_cm}
\end{figure}

\begin{figure}[!h]
    \centering
    \includegraphics[width=0.75\linewidth]{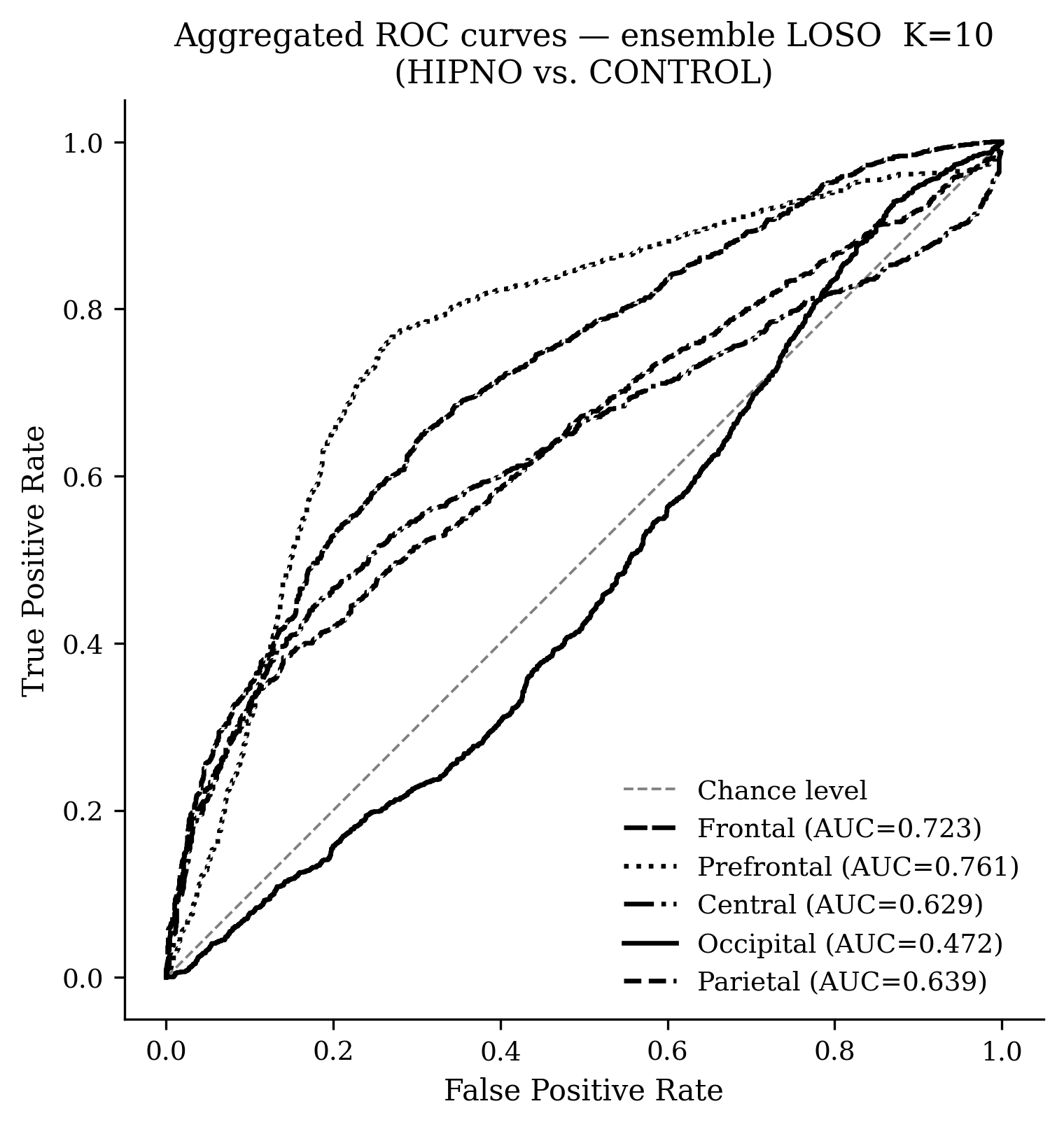}
    \caption{Aggregated ROC curves from ensemble LOSO cross-validation (K~=~10), computed
    from concatenated fold predictions. AUC values are shown in the legend.}
    \label{fig:ensemble_loso_roc}
\end{figure}

An ensemble variant of LOSO cross-validation was applied to examine whether aggregating
predictions across multiple temporal windows improves subject-level generalisation,
as described in Materials and Methods. Two ensemble sizes were evaluated: K~=~3 and K~=~10. Table~\ref{tab:ensemble_loso} reports
results for K~=~10 alongside the single-window LOSO baseline; Figures~\ref{fig:ensemble_loso_accuracy},
\ref{fig:ensemble_loso_cm}, and~\ref{fig:ensemble_loso_roc} present the corresponding
per-subject accuracy distributions, confusion matrices, and ROC curves.

\begin{table}[!h]
\centering
\caption{Ensemble LOSO results with K~=~10 windows (soft voting). Single-window LOSO accuracy
is included as reference. $\Delta$~ACC is relative to the single-window baseline. ACC is
mean~$\pm$~std across 62 folds; F1, Precision, Recall, and AUC are computed on concatenated
fold predictions (HIPNO as positive class).}
\label{tab:ensemble_loso}
\small\setlength{\tabcolsep}{4pt}
\begin{tabular}{lccccccc}
\hline
\textbf{Region} & \textbf{Single ACC} & \textbf{Mean ACC $\pm$ Std} & \textbf{$\Delta$ ACC} & \textbf{F1} & \textbf{Precision} & \textbf{Recall} & \textbf{AUC} \\
\hline
Prefrontal & 0.624 & $0.738 \pm 0.361$ & $+$0.114 & 0.722 & 0.744 & 0.701 & 0.761 \\
Frontal    & 0.646 & $0.667 \pm 0.347$ & $+$0.021 & 0.635 & 0.675 & 0.600 & 0.723 \\
Central    & 0.591 & $0.629 \pm 0.338$ & $+$0.038 & 0.584 & 0.637 & 0.539 & 0.629 \\
Parietal   & 0.631 & $0.609 \pm 0.352$ & $-$0.022 & 0.564 & 0.612 & 0.522 & 0.639 \\
Occipital  & 0.453 & $0.462 \pm 0.352$ & $+$0.009 & 0.429 & 0.441 & 0.417 & 0.472 \\
\hline
\end{tabular}
\end{table}

The ensemble approach yielded the largest gain for the prefrontal region, which improved from
a single-window mean accuracy of 0.624 to 0.738 with K~=~10 ($\Delta$~ACC~=~$+$0.114).
Frontal and central regions showed smaller but positive improvements of $+$0.021 and $+$0.038,
respectively. In contrast, the parietal region showed no benefit from ensembling
($\Delta$~ACC~=~$-$0.022), and the occipital region remained near chance (AUC~=~0.472)
regardless of ensemble size, confirming that aggregating windows cannot compensate for the
absence of a generalisable discriminative signal.

Comparing K~=~3 and K~=~10 reveals diminishing returns beyond the smaller ensemble size.
The prefrontal region gained only marginally from K~=~3 to K~=~10 ($+$0.007 in ACC), while
frontal, central, and parietal regions performed better with K~=~3 than with K~=~10, suggesting
that including additional, less discriminative windows introduces noise into the soft-vote
average. The high per-subject standard deviation (0.34--0.36 across all regions) reflects a
bimodal distribution of per-fold outcomes, where individual subjects are either classified with
high accuracy or near chance, consistent with the heterogeneity of individual EEG signatures
under a subject-independent evaluation scheme.

\section{Discussion}

There are well-established approaches to applying Convolutional Neural Networks in biomedical
signal classification~\cite{cnnAppOh}, and deep learning methods are increasingly adopted in
systems aimed at supporting clinical and therapeutic practice~\cite{zemla2023-glm}. The aim of
this study was to verify whether a minimally preprocessed EEG signal, without manual feature
extraction, is sufficient to classify hypnotic susceptibility across distinct brain regions using
a 2D CNN in a sliding-window framework.

The results reveal a clear posterior-anterior gradient in classification performance. Training
and evaluating up to 868 independent models across seven brain regions demonstrated that the
occipital and parietal regions provide the most consistent group separation, while anterior
regions --- though occasionally reaching competitive peak accuracies --- exhibit substantially
lower temporal stability of the discriminative signal.

The dominance of the occipital region (90.3\% of windows above threshold, mean ACC 0.7206) is
consistent with the role of the visual cortex in hypnosis-related processing. Guided Imagery,
by its nature, engages mental visual imagery, and it is well established that occipital alpha
oscillations are modulated during hypnotic induction and imagery-based tasks~\cite{graffin1995eeg,jensen2015mechanisms}. The strong
and temporally stable discriminability in this region suggests that hypnosis-related neural
signatures are not confined to brief moments of induction but persist throughout much of the
20-minute session, supporting the notion of a sustained altered state rather than a transient
response. The parietal region, which ranked second in both proportion of discriminative windows
(76.6\%) and mean accuracy (0.7087), is anatomically and functionally linked to attentional
reorientation and multisensory integration --- cognitive processes known to be modulated during
hypnosis~\cite{rainville2002hypnosis,landry2017brain}. Together, these posterior regions appear to carry the most reliable neural
markers of hypnotic susceptibility at the level of 10-second EEG epochs.

The prefrontal region presented a contrasting profile: the highest single-window accuracy of the
entire experiment (0.8224) but only 19.4\% of windows above threshold. This dissociation between
peak and temporal consistency suggests that frontal discriminability is transient, potentially
reflecting episodic engagement of top-down executive control during specific moments of the
recording rather than a sustained hypnotic state marker~\cite{egner2005hypnosis}. Similar transient frontal effects
have been reported in EEG studies of meditation and altered states of consciousness, where
prefrontal activity fluctuates in relation to depth of relaxation~\cite{lazar2000functional}. The central and frontal
regions showed analogous patterns of low window proportions combined with moderate peak
accuracies, further supporting the interpretation that anterior cortical responses to hypnosis
are time-locked to specific procedural events rather than constituting a stable baseline
difference between groups.

The complete absence of discriminative windows in the temporal and facial regions warrants
attention. Despite both groups listening to audio recordings throughout the session --- differing
only in content (relaxation versus cognitive task) --- no systematic EEG difference was detected
at the temporal level. This suggests that the neural distinction between the HIPNO and CONTROL
conditions is not primarily driven by differential auditory processing but rather by higher-order
cognitive and imagery-related mechanisms. The facial region, predominantly reflecting peripheral
muscle activity rather than cortical dynamics, was expected to contribute little to group
classification, and the results confirm this expectation.

The classification performance achieved by the 2D CNN is broadly comparable to results reported
for classical ML approaches in similar paradigms. SVM-based classifiers applied to hypnosis and
relaxation-related EEG have achieved accuracies of approximately 80\%~\cite{svmRelaxation}, and
the GLM approach used in our previous work~\cite{zemla2023-glm} attained 81\% on a selected
60-second time segment using handcrafted frequency-band features. The present framework matches
and in some windows exceeds these benchmarks without requiring manual feature extraction,
operating instead directly on bandpass-filtered and z-score-normalised raw signal. This supports
the argument that CNN-based end-to-end learning is a viable alternative to feature-engineered
pipelines for EEG-based state classification. Simple filtering and z-score normalisation of the
signal appear to be sufficient for achieving classification performance comparable to that of
state-of-the-art methods.

The strict subject-independent evaluation provided by LOSO cross-validation substantially
revises the spatial picture suggested by the random-split results. The occipital region, which
achieved the highest random-split accuracy (mean ACC 0.7206) and discriminated 90.3\% of windows
above threshold, collapsed to near-chance performance under LOSO (mean ACC 0.453, AUC 0.429 ---
below the 0.5 chance level). This reversal indicates that the high random-split accuracy was
driven by subject-specific EEG fingerprints rather than genuine group-level differences related
to hypnotic susceptibility. EEG signals are known to contain highly individual spectral and
spatial characteristics that can serve as biometric identifiers~\cite{zhang2022eeg}; when
segments from the same recording session appear in both training and test sets, a classifier
can learn to identify individuals rather than groups, artificially inflating apparent accuracy.
The occipital region is particularly susceptible to this effect given the high within-session
stability of posterior alpha rhythms.

In contrast, the frontal and prefrontal regions --- which showed lower and less consistent
discriminability in the random-split analysis --- maintained moderate, above-chance performance
under LOSO: frontal (mean ACC 0.646, AUC 0.692), prefrontal (0.624, AUC 0.681), and parietal
(0.631, AUC 0.680). This suggests that anterior and parietal EEG features carry more genuinely
generalisable neural signatures of the HIPNO/CONTROL distinction than the random-split results
implied. The inversion of the posterior-anterior ranking under cross-subject evaluation underscores
the necessity of subject-independent validation in EEG-based classification studies.

The ensemble LOSO experiments further demonstrate that aggregating predictions across multiple
temporal windows most benefits the prefrontal region, which improved from 0.624 to 0.738 with
K~=~10 windows ($\Delta$~ACC~=~$+$0.114). This finding aligns with the interpretation that
prefrontal hypnosis-related activity is episodic and temporally heterogeneous: no single
10-second window reliably captures the discriminative signal, but combining predictions from
multiple windows reduces variance and recovers a more stable classification boundary. The
diminishing returns from K~=~3 to K~=~10 for most regions suggest that an ensemble of three
windows already captures the most informative temporal variation for frontal, central, and
parietal regions, and that including additional windows of lower discriminability introduces
noise rather than signal.
The bimodal distribution of per-subject outcomes --- where individual subjects are classified
with either high accuracy or near chance --- may additionally reflect genuine inter-individual
heterogeneity in the depth of GI-induced relaxation: subjects who engage deeply with the imagery
task are likely to produce more pronounced and temporally stable anterior EEG signatures, whereas
subjects who do not reliably enter the relaxed state would be expected to yield near-chance
classification under a group-level model. This interpretation is consistent with the
well-documented variability in hypnotic susceptibility across individuals and represents an
inherent challenge for any EEG-based mental-state classifier trained on binary group labels.

The main limitation of this work is the relatively small number of subjects available for
training. The initial train-test split (90\%/10\%) was applied at the segment level rather than
at the subject level, allowing signal from the same individual to appear in both training and
test sets and inflating accuracy estimates --- an effect confirmed by the occipital results
collapsing to below chance under LOSO. This limitation was addressed by the LOSO and ensemble
LOSO experiments reported above, which provide a subject-independent estimate of generalisation
performance. The study also recruited exclusively male participants aged
19--24, limiting the generalisability of the findings; prior research has documented sex-related
differences in EEG spectral characteristics~\cite{wada1994gender,cantillo2017gender}, and it
remains to be established whether the observed spatial patterns of discriminability replicate in
female cohorts.
More broadly, the cohort consisted of healthy volunteers with no history of chronic illness
or medication use; the extent to which a model trained on this population generalises to clinical
groups --- including patients with comorbid conditions, psychiatric disorders, elderly individuals,
or those undergoing pharmacological treatment --- remains an open and important question for
translational validation.
No artifact rejection or independent component analysis (ICA) was performed,
meaning that residual ocular and muscle artifacts may have contributed to classification
performance in certain electrode regions --- although the absence of discriminative windows in
the facial region suggests that this factor did not introduce systematic group differences.
In particular, frontal and prefrontal electrode subsets overlap spatially with peri-orbital
artefact sources; although all participants performed the tasks with eyes closed in a darkened
room --- substantially reducing eye-blink artefacts --- residual slow ocular movements and
electrode drift near the orbit cannot be fully excluded as contributors to the anterior
classification signal without ICA-based source separation.
There is still room for improving these models by collecting data from more subjects and testing
more complex architectures. Additionally, investigating alternative electrode subsets --- for
instance based on the 10--20 international system~\cite{chatrian1985ten} --- could inform
whether reduced-channel configurations retain the discriminative power of the full
256-electrode setup.
The 256-channel EGI dense array system requires trained personnel, conductive gel, and
extended setup time, which limits its practicality in frontline clinical environments such as
oncology day wards, pain clinics, or outpatient psychology departments; future work should
assess whether reduced-channel or consumer-grade EEG devices can achieve comparable
classification performance.
Furthermore, all participants in this study were drug-naive, whereas clinical patients are
frequently receiving pharmacological treatments --- including opioids, benzodiazepines,
antidepressants, or chemotherapy agents --- many of which are known to alter EEG spectral
characteristics. Whether the classifier's performance is maintained in medicated patient
populations requires dedicated investigation.

A further conceptual limitation concerns the asymmetry between experimental conditions.
Although both groups performed the tasks with eyes closed while listening to audio recordings
--- controlling for basic sensory input --- the GI group engaged in a passive, self-referential
relaxation task while the Control group performed active memorisation under time pressure.
The classifier therefore operates on a composite contrast encompassing differences in cognitive
load, task engagement, and attentional focus, in addition to any effects specific to hypnotic
state induction. Disentangling the contribution of hypnotic suggestibility from the broader
relaxation-versus-cognitive-effort contrast would require an additional control condition ---
such as passive listening to a non-GI audio recording --- that matches the GI condition in
sensory input while withholding the induction element. This represents an important direction
for future experimental work.
It should also be noted that the study classifies EEG \emph{states} during GI exposure
versus cognitive task performance, rather than \emph{trait} hypnotic susceptibility. Future
work should examine whether the EEG features identified here correlate with standardised
susceptibility scores, which would be a necessary step toward predicting individual treatment
response.

\section{Conclusions}

It was demonstrated that a 2D Convolutional Neural Network trained on minimally preprocessed EEG
signal can discriminate between subjects with high hypnotic susceptibility and control subjects
with accuracy exceeding the 0.67 threshold in five out of seven evaluated brain regions. The
occipital and parietal regions emerged as the most reliable sources of discriminative neural
information, exhibiting sustained group separation across the majority of the 20-minute recording.
The prefrontal region, while yielding the highest peak accuracy (0.8224), showed limited temporal
stability, indicating transient rather than sustained frontal engagement. Temporal and facial
regions provided no discriminative signal within the applied framework.

The results show that simple filtering and normalisation of the raw signal, combined with a
lightweight 2D CNN, are sufficient to achieve classification performance comparable to classical
machine learning methods employing handcrafted frequency-band features. The sliding-window
approach proved effective in identifying temporally localised periods of maximal group
separability, offering a data-driven alternative to expert-defined time-segment selection.

Research using electroencephalography to study hypnotic susceptibility could significantly
contribute to the development of brain-computer interfaces tailored for therapeutic applications.
By analysing EEG patterns associated with distinct neurological states induced by Guided Imagery,
researchers can identify specific biomarkers indicative of a hypnosis-prone state --- such as
alterations in occipital and parietal brainwave activity commonly linked to relaxation and
heightened imaginative engagement. Integrating these EEG-derived biomarkers into a real-time BCI
system could allow therapists to receive objective feedback on a patient's depth of immersion in
the Guided Imagery process, enabling them to optimise the timing and effectiveness of therapeutic
interventions.

So, as shown above, the research presented here can shed new light on the engineering of new
brain-computer interfaces with application for psycho-therapists and neuro-therapists using
relaxation techniques and the Guided Imagery method.

\section*{Author Contributions}
Conceptualization, K.W.\ and G.M.W.; methodology, K.W.\ and F.P.; software, K.W.; validation,
K.W., F.P.\ and A.K.; formal analysis, K.W.; investigation, K.Z.\ and G.S.; resources, K.Z.;
data curation, K.W.; writing---original draft preparation, K.W.; writing---review and editing,
G.M.W., F.P., A.K., K.Z.\ and G.S.; visualization, K.W.; supervision, G.M.W.; project
administration, G.M.W.; funding acquisition, G.M.W. All authors have read and agreed to the
current version of the manuscript.

\section*{Institutional Review Board Statement}
The studies involving human participants were reviewed and approved by the Maria
Curie-Sk\l{}odowska University Bioethical Commission (approval code: 76/WSP/20, approval date:
09/07/2021). Data collection was conducted between 15/09/2021 and 30/11/2021.

\section*{Informed Consent Statement}
The participants provided their written informed consent to participate in this study. Informed
consent was obtained from all subjects involved in the study.

\section*{Data Availability Statement}
The datasets generated and analysed during the current study are available from the
corresponding author on reasonable request. All analysis code is available in the project
repository.

\section*{Acknowledgments}
The authors thank all participants for their time and cooperation. During the preparation of
this manuscript, the author(s) used GPT-5.6 Luna for minor editorial support. The authors have
reviewed and edited the output and take full responsibility for the content of this publication.

\section*{Conflicts of Interest}
The authors declare no conflicts of interest.

\bibliographystyle{unsrtnat}
\bibliography{references}

\end{document}